\pdfoutput=1

\documentclass[11pt]{article}

\usepackage{acl}

\usepackage{times}
\usepackage{latexsym}
\usepackage{hyperref}
\usepackage{url}
\usepackage{graphicx}
\usepackage{tabularx}
\usepackage{booktabs}
\usepackage{multirow}

\usepackage[T1]{fontenc}

\usepackage[utf8]{inputenc}

\usepackage{microtype}

\usepackage{inconsolata}

\title{Reformulating Domain Adaptation of Large Language Models as Adapt-Retrieve-Revise: A Case Study on Chinese Legal Domain}

  \author{Zhen Wan \\ Kyoto University \\ \texttt{zhenwan.nlp@gmail.com} 
  \And
  Yating Zhang \\ Alibaba Health \\ \texttt{yatingz89@gmail.com} 
  \And
  Yexiang Wang \\ Alibaba Quark \\ \texttt{yexiang\_w@163.com} 
  \AND
  Fei Cheng \\ Kyoto Univeristy \\ \texttt{feicheng@i.kyoto-u.ac.jp} 
  \And
  Sadao Kurohashi \\ Kyoto University \\ \texttt{kuro@i.kyoto-u.ac.jp} }

\begin{document}
\maketitle
\begin{abstract}
While large language models (LLMs) like GPT-4 have recently demonstrated astonishing zero-shot capabilities in general domain tasks, they often generate content with hallucinations in specific domains such as Chinese law, hindering their application in these areas. This is typically due to the absence of training data that encompasses such a specific domain, preventing GPT-4 from acquiring in-domain knowledge. A pressing challenge is that it's not plausible to continue training LLMs of the GPT-4's scale on in-domain data.

This paper introduces a simple yet effective domain adaptation framework for GPT-4 by reformulating generation as an \textbf{adapt-retrieve-revise} process. The initial step is to \textbf{adapt} an affordable 7B LLM to the Chinese legal domain by continuing learning in-domain data. When solving an in-domain task, we leverage the adapted LLM to generate a draft answer given a task query. Then, the draft answer will be used to \textbf{retrieve} supporting evidence candidates from an external in-domain knowledge base. Finally, the draft answer and retrieved evidence are concatenated into a whole prompt to let GPT-4 assess the evidence and \textbf{revise} the draft answer to generate the final answer. 

Our proposal combines the advantages of the efficiency of adapting a smaller 7B model with the evidence-assessing capability of GPT-4 and effectively prevents GPT-4 from generating hallucinatory content. In the zero-shot setting of four Chinese legal tasks, our method improves the average score by $+33.6$ points, compared to GPT-4 direct generation. When compared to two stronger retrieval-based baselines, our method outperforms them by $+17.0$ and $+23.5$. The code for training our model is here:~\footnote{ \url{https://github.com/YukinoWan/Adapt-Retrive-Revise}}.
\end{abstract}

\section{Introduction}
\begin{figure}[t!]
    \centering
    \scalebox{1.0}{
    \includegraphics[width=1.0\linewidth]{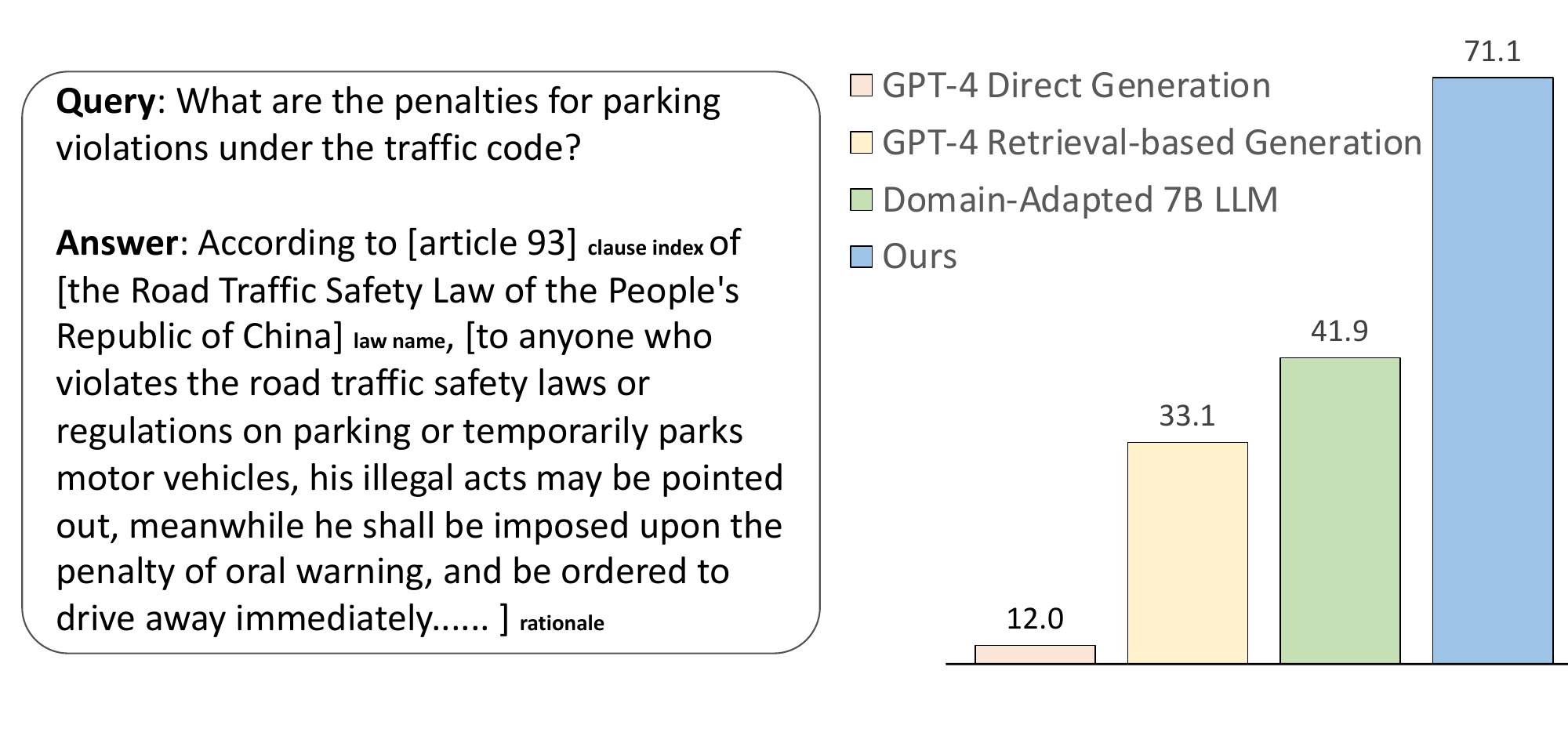}}
    \caption{\textbf{Left}: A real translated example of Chinese LegalQA. The square brackets and subscripts are offered for the purpose of clear demonstration, not actually exist in the ground-truth answer or generation.  \textbf{Right}: Models' F1 scores on the LegalQA dataset.}
    \label{fig:abstract}
\end{figure}
Recent large language models like GPT-4 bring remarkable improvements in various general domain NLP tasks \citep{NEURIPS2020_1457c0d6,openai2023gpt4,thoppilan2022lamda,chowdhery2022palm,rae2022scaling,hoffmann2022training}.
However, in specific domains such as the Chinese legal domain, the performance of such general LLMs still lags considerably behind. We show a real example of Chinese LegalQA~\citep{DBLP:conf/icail/ChenYZZSLS23} on the left of Figure~\ref{fig:abstract}, which requires the model to generate the corresponding legal provision (i.e., the law name and the clause index) and the rationale for the judgment, given a brief case description as the query.

\begin{figure}[t!]
    \centering
    \scalebox{1.0}{
    \includegraphics[width=1.0\linewidth]{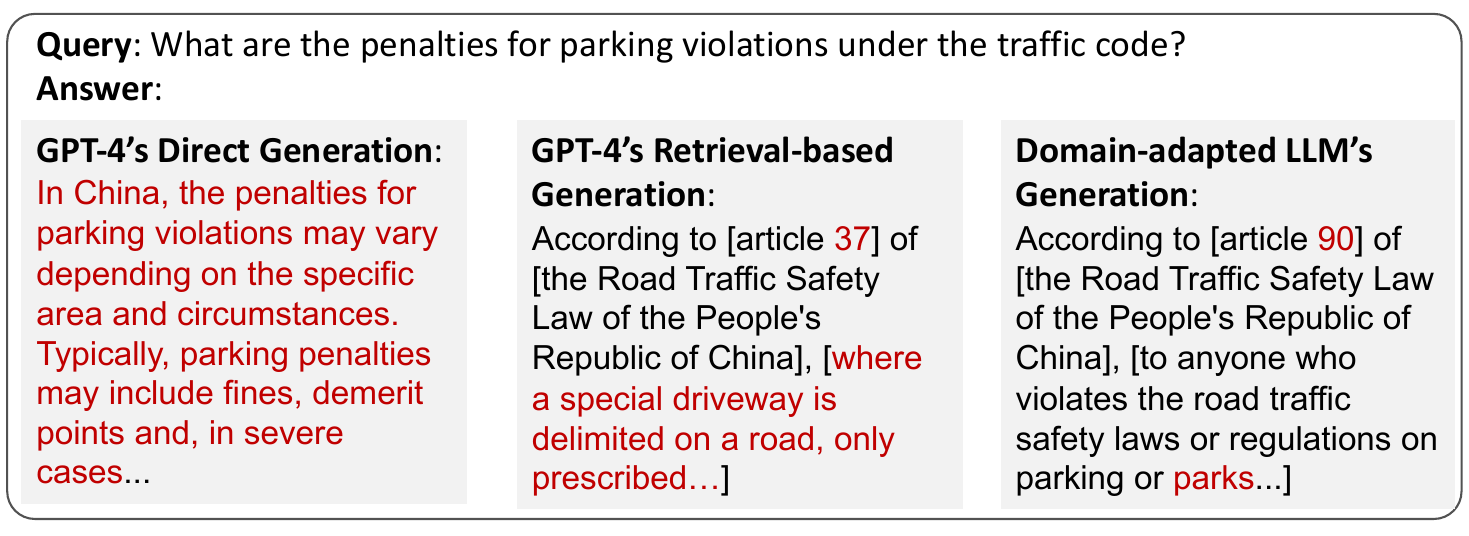}}
    \caption{Examples of hallucinations of various models. \textcolor{red}{Red} denotes the content containing hallucinations. The ground-truth answer refers to the left case in Figure~\ref{fig:abstract}.}
    \label{fig:examples}
\end{figure}

We initialize the research with a preliminary examination of utilizing GPT-4 to address the Chinese LegalQA task, which involves responding with a law clause relevant to a given query case. Figure~\ref{fig:abstract} reveals the extremely low performance (F1 $12.0$) of directly prompting the query case to ask GPT-4 to generate the corresponding law clause. Though the generated answers are grammatically fluent, they often consist of non-logical content, factual mistakes, and fail to refer to the correct legal provision (also known as "hallucination"). For example, in the left answer of Figure~\ref{fig:examples}, the direct generation of GPT-4 seems to retell the case description but fails to point out the corresponding clause. A potential reason is the insufficient Chinese legal domain text during pretraining, leading to a lack of domain knowledge acquisition, and consequently generating hallucinatory content. 

For the LLMs with the scale of GPT-4, it's generally not feasible for researchers to do learning-based adaptation. The enormous model size could make the cost of continual learning extremely high, and meanwhile, the access functions are often limited by APIs. Therefore, recent work \citep{DBLP:conf/nips/LewisPPPKGKLYR020,DBLP:conf/iclr/0002IWXJ000023,shuster2021retrieval,ma2023query} introduces retrieval-based methods that first use the given query to retrieve relevant evidence candidates from the external domain-specific knowledge base or the internet and then concatenate the query and the evidence candidates into the prompt. GPT-4 could implicitly validate the relevance between the query and the evidence, as well as the correctness of the evidence, before producing a generation.

Our replicated retrieval-based method improves the LegalQA F1 from the $12.0$ points of direct generation to $33.1$ as shown in Figure~\ref{fig:abstract}. It indicates that even though GPT-4 may not generate domain content, it possesses sufficient evidence-assessing capacity to select the correct evidence from candidates.
Nevertheless, the retrieval module is limited by the capability of representation mapping from query to evidence and is also influenced by the domain issue, leading to a decline in search quality. GPT-4 still produces hallucinations in responses as the middle answer in Figure~\ref{fig:examples}. 

On the other hand, with the rapid development of open LLMs led by LLaMA \citep{DBLP:journals/corr/abs-2302-13971}, it becomes affordable to continually train an open LLM tailored to your demands on sufficient in-domain texts, resulting in a domain-adapted LLM. We therefore conduct the second examing of continually training Baichuan 7B~\citep{Baichuan-7B}, a Chinese foundation model, on over 50B token Chinese legal data. Its performance (the green bar in Figure~\ref{fig:abstract}) even surpasses the retrieval-based GPT-4 generation on Chinese LegalQA. Hallucinations caused by the lack of domain knowledge are largely reduced but not completely solved. As shown in the right answer in Figure~\ref{fig:examples}, adapted LLM generates generally correct responses but still makes errors in certain words. Although the law name is correct and the rationale part is reasonable, the clause index is a hallucinatory generation which raises the difficulty in anchoring target clauses. We argue that these fails are accordingly caused by the limited capability of a 7B size to memorize the knowledge accurately.

Building upon the observation of the evidence-assessing capability of GPT-4 and the high-quality domain content generated by the domain-adapted 7B model, this paper proposes a novel approach to reformulate GPT-4's domain content generation to an \textbf{adapt-retrieve-revise} process: (1) the \textbf{domain-adapted} model generates a draft answer given a query; (2) the \textbf{retrieval} module uses the draft answer as input for searching external evidence candidates because the answer is usually more informative and semantically similar to the evidence compared to the query as long as the answer quality is acceptable; (3) GPT-4 assesses retrieved evidence and \textbf{revises} the draft answer to generate the final answer.

The rest sections of the paper anchor the Chinese legal domain and comprehensively validate the effectiveness of our proposal. In Section~\ref{sec:method}, we explain each component of our adapt-retrieve-revise method and elaborate on the implementation details. In Section~\ref{sec:exp} and~\ref{sec:abla}, we conduct the experiments and the result analysis on four Chinese legal domain tasks. The experimental results show substantial improvements against the direct generation and the retrieval-based generation baselines. In the final Section~\ref{sec:con}, we elicit the conclusion and future work. To the best of our knowledge, this is the first study to examine the zero-shot performance of LLMs on four Chinese legal benchmarks. 

\begin{figure*}[t]
    \centering
    \includegraphics[width=\linewidth]{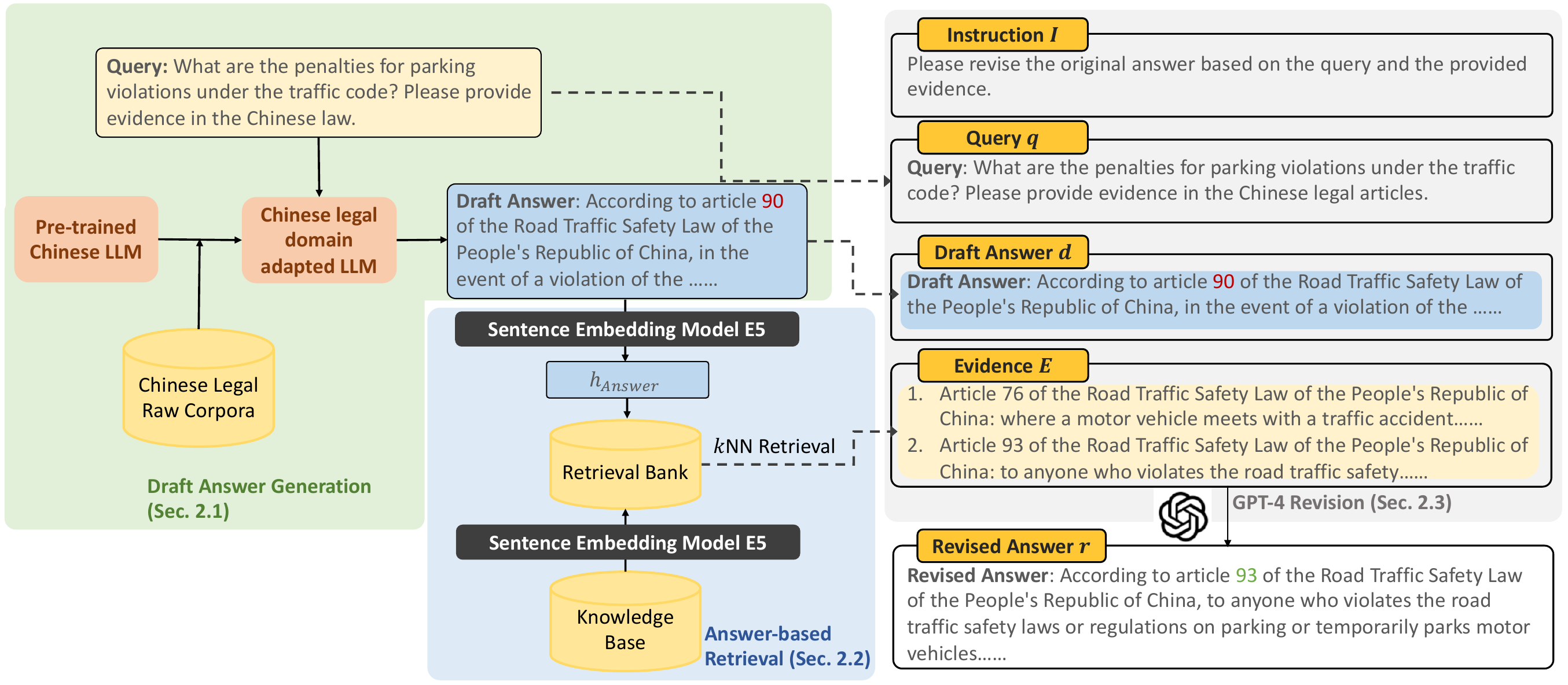}
    \caption{\textbf{Overview of our proposed method.} The example and prompt are translated from Chinese to English for the demonstration purpose.}
    \label{fig:overview}
\end{figure*}

\section{Methodology}
\label{sec:method}
Our adapt-retrieve-revise method consists of three steps. In the first step (Section~\ref{step1}), we continually train a Chinese pre-trained LLM on the Chinese legal domain corpora to derive a domain-adapted legal LLM and given the query, the legal LLM will generate the draft answer. In the second step (Section~\ref{step2}), we use a sentence embedding model to produce embeddings for both the draft answer and each paragraph in the corresponding knowledge base, then evidence retrieval will be computed by the similarities between the answer embedding and the paragraph embeddings. In the third step (Section~\ref{step3}), we concatenate the query, the draft answer, and the retrieved evidence in the prompt for GPT-4 to revise and produce the final response. Figure~\ref{fig:overview} shows the overview of our method. In the following sections, we will introduce details of each step.

\subsection{Draft Answer Generation by the Domain-adapted LLM}\label{step1}
This step could be actually flexible to save the effort of training a LLM yourself. For instance, an off-the-shelf Chinese legal domain LLM can be used for generating draft answer. However, most of such LLMs have been fine-tuned on various in-domain evaluation tasks. The potential data leakage could lead to unreliable evaluation, especially in a zero-shot setting. Therefore, in this paper, we adapt an open 7B LLM to the Chinese legal domain by continual learning on the domain data.
We collect the training data from the following two open Chinese legal sources:
\begin{itemize}
    \item \textbf{Chinese Law Clauses} (\url{https://flk.npc.gov.cn/}) form the foundation of the judicial system, containing a wealth of legal terms, provisions, and judicial practices. They are essential for the model to understand and generate relevant content.
    \item \textbf{Chinese Judgments Online} (\url{https://wenshu.court.gov.cn/}) is the largest online publication platform for legal documents in China. The platform contains judicial documents from courts at all levels, covering various legal fields such as civil, criminal, administrative, and enforcement. Such documents contain knowledge for LLMs to understand the usage of laws in various scenarios.
\end{itemize}

During the inference, given an input query, we will first prompt the trained 7B legal LLM to generate the draft answer, which will be used in the next step. For the prompt, we add the instruction ''Please provide evidence in the Chinese law" at the end of the query to enforce the model to generate related law clauses, as in Figure~\ref{fig:overview}.

\subsection{Answer-based Evidence Retrieval}\label{step2}
Since the draft answer of the 7B legal LLM is usually more informative and semantically similar to the evidence than the query. We further use the generated evidence to retrieve ground-truth evidence from the target knowledge base for the purpose of revision since it contains much more information than the query, even though the hallucinations can not be totally reduced. We implement this method with two subsequent steps: knowledge bank construction and retrieval.
\paragraph{Knowledge Bank Construction}
For the $i$-th paragraph $p_i$, we construct the \textit{key-value} pair $(\mathbf{p_i}, p_i)$ where the \textit{key} $ \mathbf{p_i}$ is the representation obtained from the sentence embedding model and the \textit{value} $p_i$ denotes the paragraph. The memory $(\mathcal{K}, \mathcal{V})=\{(\mathbf{p_i}, p_i)|p_i\in \mathcal{KB}\} $ is thus the set of all \textit{key-value} pairs constructed from all the paragraphs in the external knowledge base $\mathcal{KB}$.

\paragraph{Retrieval}
Given the generated draft answer $d$, the sentence embedding model outputs its representation $\mathbf{h_{Answer}}$. We then query the constructed knowledge bank with $\mathbf{h_{Answer}}$ to retrieve its $k$ nearest neighbors $E$ according to a distance function by $L^{2}$ distance.

\subsection{GPT-4 Revision}\label{step3}
To effectively combine the high-quality draft answers generated by the 7B domain adapted model with GPT-4's powerful evidence-assessing capability, we propose the following process.
As shown in Figure~\ref{fig:overview}, the whole prompt consists of the following components: (1) the instruction $I$ to require GPT-4 to revise the draft answer given the query and the evidence candidates; (2) the query $q$ itself; (3) the draft answer $d$ for GPT-4 to revise; (4) and the retrieved evidence candidates $E$ to provide related Chinese legal knowledge for GPT-4. Then, the final revised answer $r$ will be outputted by
$GPT4( I,q, d, E )$.

\section{Experiment Settings}
\label{sec:exp}
We conducted a series of experiments to compare our adapt-retrieve-revise method to the baselines of direct generation and retrieval-based generation on various Chinese legal benchmarks. 
We show the model details and the task settings in this section.
\subsection{Model Settings}
\paragraph{Details of training 7B legal LLM:} We utilize the general domain Baichuan 7B model\footnote{\url{https://huggingface.co/baichuan-inc/Baichuan-7B}} for continual learning Chinese legal corpora. In total, we trained 50B tokens of Chinese Law Clauses and Chinese Judgments Online corpora with the input length limit of 16K and the batch size of 256 on 32 A100 GPUs, and the time-consuming is 167 hours.
After continual learning, we subsequently supervised fine-tuning our model on 70$K$ instruction examples, including 52$K$ GPT-4 self-instruct Chinese data~\citep{peng2023instruction} and 18K legal instructions (See Appendix~\ref{legal sft}) for the alignment.

\paragraph{Retriever setting:} We utilize Multilingual-E5-large \citep{DBLP:journals/corr/abs-2212-03533}, a Roberta-based~\citep{liu2019roberta} sentence embedding model that achieves robust performance on various tasks. We also compare with other retrieval modules in Appendix~\ref{retrieval comparison}.

\paragraph{GPT-4 setting:} For the utilization of GPT-4, we select ``gpt-4-0613'' with maximal 8$K$ input tokens and use the original Chinese prompt as shown in Sec~\ref{step3} and Figure~\ref{fig:overview} via OpenAI API.

\subsection{Evaluation of Chinese Legal Tasks}
We evaluated our Adapt-Retrieve-Revise method on a diversity of tasks with different knowledge base for retrieval in the zero-shot setting:
\begin{itemize}
  \item \textbf{Law Clause Recommendation (LCR) and Criminal Prediction (CP)} \citep{DBLP:journals/corr/abs-1807-02478} are two tasks using the legal report as the input, and let the model generate the most related law clause and predict the criminal type based on the law clause. For these two tasks, we use the \textbf{Chinese law clauses} as the domain knowledge base for retrieval.
  \item \textbf{LegalQA} is a filtered set of EUQALS~\citep{DBLP:conf/icail/ChenYZZSLS23} that, given an input query, the model should generate an answer based on the most related legal clause. The filtering is based on the quality of the questions and we will release the filtered set. We also use the \textbf{Chinese law clauses} as the domain knowledge base for retrieval.
  \item \textbf{JEC-QA} \citep{DBLP:conf/aaai/ZhongXTZ0S20} is the official test for getting a lawyer's certificate in China. We chose the single-choice selection questions in our evaluations with the \textbf{Legal Textbooks} (\url{https://github.com/thunlp/jec-qa}) as the knowledge base for retrieval.
  \item \textbf{Similar Case Retrieval}~\citep{DBLP:conf/sigir/MaSW000M21} is the task, given a query legal scenario as the input, we aim at selecting similar \textbf{Legal Judgement Documents} from the $100$ candidates. We conducted this experiment to assess the reliability of our proposed retrieval method in Section~\ref{similar-case-retrieval}.

\end{itemize}

\begin{table*}[t!]
\centering
    
    \resizebox{\linewidth}{!}{
    \begin{tabular}{l|l|l|c|c|c|c|c}
    \toprule
        \multirow{3}{*}{Generator} & \multirow{3}{*}{Retriever} & \multirow{3}{*}{Revisor} &\multicolumn{3}{c}{Chinese law clauses} & Textbooks& \multirow{3}{*}{Avg.}\\
        \cline{4-7}
         &  &  &LCR& CP& LegalQA&JEC-QA& \\
         &  &  &F1 (Rec.)& F1 (Rec.)& F1 (Rec.)&Acc.& \\
         \toprule
         \multicolumn{8}{c}{\textit{Direct Generation}} \\
         \hline
         GPT-4&-&- &61.7 (67.6)& 70.7 (71.2) & 12.0 (14.4) & 36.2&45.1\\
         \hline
         LawGPT-7B~\cite{LawGPT}&-&- &19.3 (25.3) & 33.3 (34.6) & 10.3 (16.6) & 27.4&22.6\\
         \hline
         ChatLaw-13B~\cite{cui2023chatlaw}&-&- &25.6 (27.6)& 43.6 (49.6) &14.6 (18.3)& 31.8&28.9\\
         \hline
         7B legal LLM&-&- &83.0 (88.4)& 82.9 (84.0) & 41.9 (48.8) & 39.8&61.9\\
         \bottomrule
         \multicolumn{8}{c}{\textit{Retrieval-based Generation}} \\
         \hline
         GPT-4 & Query-based&-&72.0 (74.4)& 74.0 (75.2) & 33.1 (36.0) & 41.6&55.2\\
         \hline
         7B legal LLM & Query-based&-&77.3 (87.6)& 81.3 (82.4) & 47.4 (50.2) & 40.8&61.7\\
         \bottomrule
         \multicolumn{8}{c}{\textit{Adapt-Retrieve-Revise}} \\
         \hline
         7B legal LLM & Answer-based&7B legal LLM&84.1 (88.4)& 82.7 (83.6) & 47.3 (49.0) &40.2&63.6\\
         \hline
         7B legal LLM (ours) &Answer-based&GPT-4&\textbf{90.6 (96.4)}& \textbf{86.9 (87.8)} & \textbf{71.1 (72.4)} & \textbf{66.2}&\textbf{78.7}\\
         \bottomrule
    \end{tabular}
    }
\caption{\textbf{Zero-shot Results on four Chinese legal datasets}. ``Rec.'' denotes recall, ``Acc.'' denotes accuracy, and ``Avg.'' is computed by the F1 and accuracy scores of all four tasks} 
\label{main results}
\end{table*}

Due to the cost of GPT-4 API and the human evaluation, we randomly sampled a subset of $250$ test examples for each task of LCR, CP, LegalQA, and JEC-QA. Please refer to Table~\ref{dataset} for the statistics.
\begin{table}
    \centering
    \resizebox{1.0\linewidth}{!}{
    \begin{tabular}{l|r|r|r|r}
    \toprule
        Datasets&CP& LCR&LegalQA & JEC-QA\\
        \hline
        \# test &965, 219& 965, 219& 1,000 & 13,341\\
        \bottomrule
    \end{tabular}
    }

\caption{\textbf{Statistics of datasets}. } 
\label{dataset}
\end{table}
\subsection{Evaluation Metrics}
Since generative models produce diverse formats in the output and the Chinese legal domain has its own features, checking the evaluation metrics in the experiments is crucial. 
For tasks LCR, CP, and LegalQA, our metric is the \textit{Micro F1} and the \textit{Recall} of whether the title of the ground-truth law clause is included in the generated answer. This is because, in real-world applications, with the correct title, the contents of the law clause can be easily revised by the rule-based system, indicating that the title is more important than the content. 

For the JEC-QA task, we use accuracy as the metric, but controlling the output into an identical format for automatic evaluation is difficult, especially for the 7B LLM that has not been fine-tuned on the JEC-QA task. We select human evaluation to ensure its accuracy by manually comparing the model output and the gold answer provided by the dataset.

For the Similar Case Retrieval task, we chose the widely used \textit{precision@k} and \textit{MAP} as the evaluation metrics.

\section{Main Experimental Results}

We provide the main results as in Table~\ref{main results}. As we claimed before, this is the first work to targeting the zero-shot LLM performance on Chinese legal domain tasks. Generally, we compare our adapt-retrieve-revise proposal with baselines of direct generations and retrieval-based generations using the query, showing that our method outperforms all baselines by substantial margins. Our main results also provide some ablation results.

We first observe the effectiveness of domain adaption. Our 7B legal LLM significantly beats GPT-4, and even compared with the retrieval-based generation of GPT-4, the 7B legal LLM still outperforms on three tasks and has competitive results on the JEC-QA task, indicating that our continual learning on Chinese legal raw corpora shows a fast and effective domain adaptation on various legal tasks. However, related work (LawGPT and ChatLaw) fails to benefit largely from the continual training due to the much less data used in training, and their base models are multilingual Llama.

Then, considering the results of GPT-4 and the GPT-4 retrieval-based generation, we find that after providing evidence of related legal knowledge, GPT-4 can improve its responses significantly ($+10.1$ points). This indicates that the retrieval-based method is a proper way to reduce hallucinations caused by the lack of domain knowledge, and owing to the robust evidence-assessing capacity, GPT-4 can adapt to the Chinese legal domain well with convincing evidence available.

\begin{figure}[t!]
    \centering
    \scalebox{1.0}{
    \includegraphics[width=1.0\linewidth]{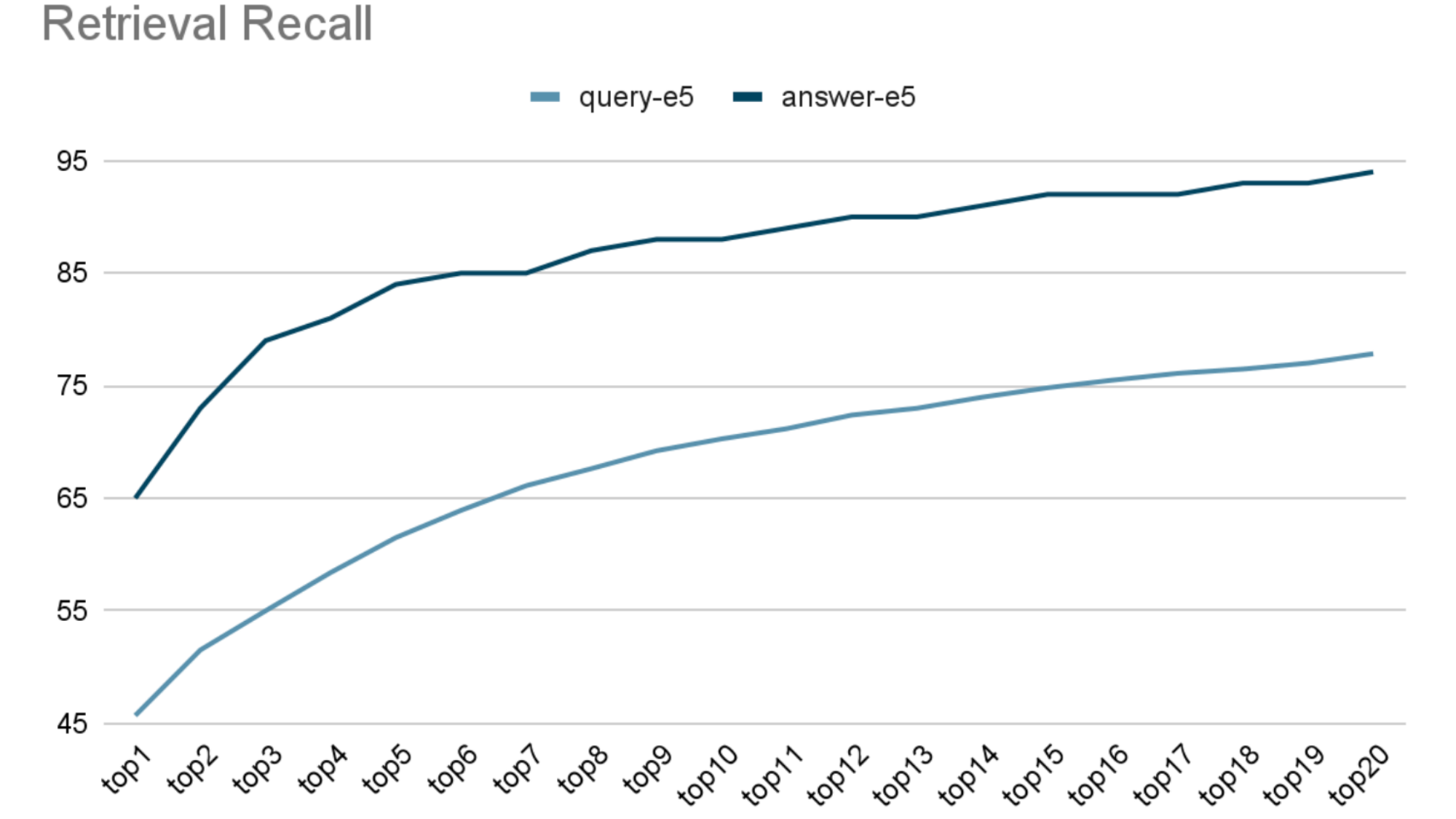}}
    \caption{\textbf{Comparison of retrieval recalls on the LegalQA dataset.}}
    \label{fig: retrieval}
\end{figure}

\begin{figure*}[t!]
    \centering
    \scalebox{0.78}{
    \includegraphics[width=1.0\linewidth]{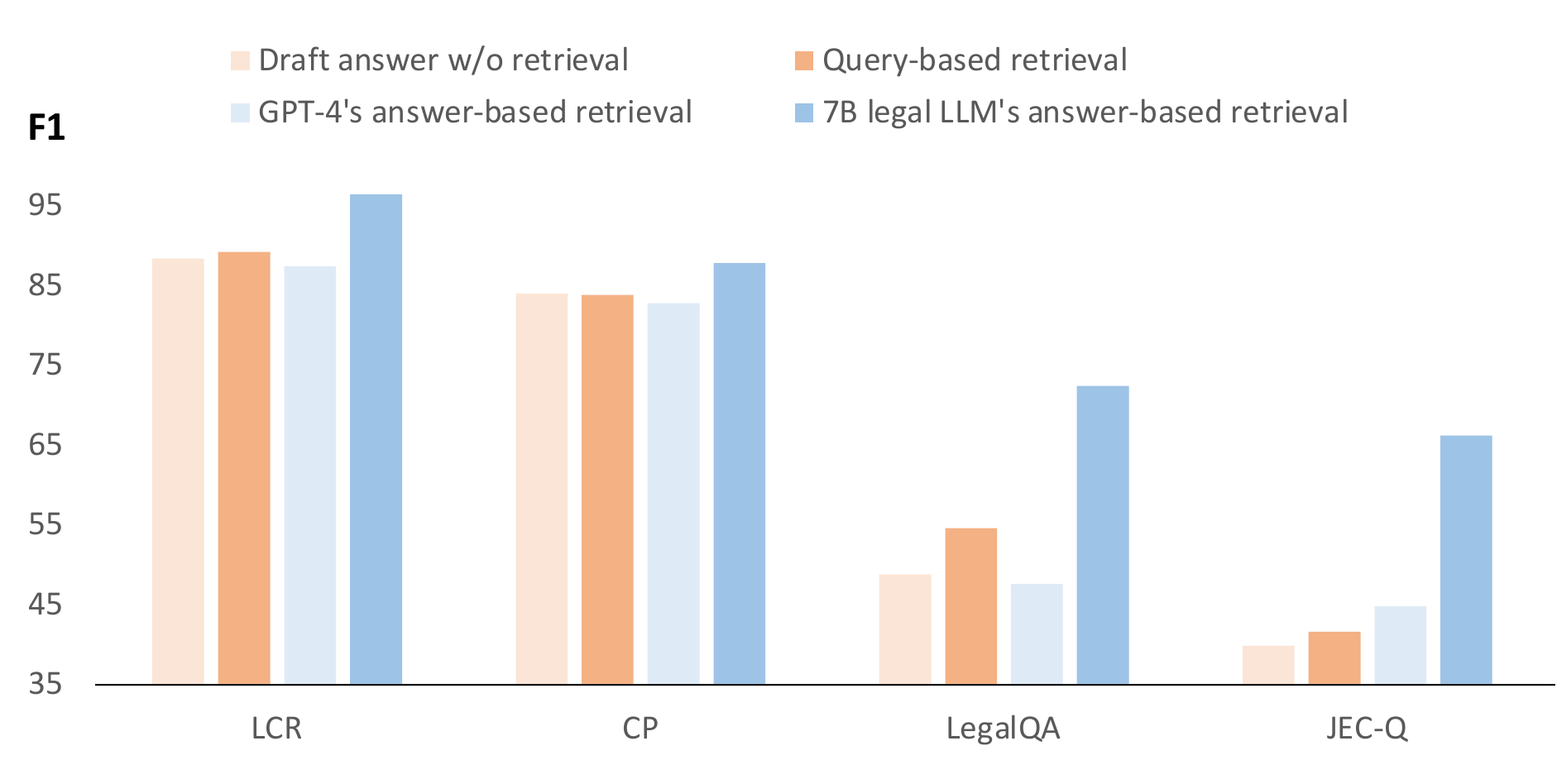}}
    \caption{We compare performances of the draft answer of 7B legal LLM and our proposed adapt-retrieve-revise model using different contents in retrieval.}
    \label{fig: retrieval-results}
\end{figure*}

In our final experiment, using the draft answers generated by the 7B legal model for retrieval and revision, the performance significantly exceeded two query-based retrieval baselines by large margins of $+17.0$ and $+23.5$ points. It's worth noting that the improvement here comes from both the enhanced answer-based retrieval quality and the revision setup. In the subsequent ablation study, we will further examine the quantified improvement of the retrieval quality through an additional retrieving task.

An interesting observation is that, by comparing the direct generation of the 7B legal model and the adapt-retrieve-revise method with the revision model as the legal 7B model, we find that with retrieved evidence, the revised answers seem to be no obvious difference from the direct generation. This indicates that the 7B legal LLM shows almost zero evidence-assessing capacity.

\section{Further Analysis}

\label{sec:abla}
\subsection{Analysis of Retrieval Methods}\label{retrieval modules}

The previous section has demonstrated the step-wise effectiveness of our adapt-retrieve-revise proposal. Nevertheless, the retriever, as a key component that directly affects the quality of evidence, how its variations would impact the final performance is a crucial research question to investigate.

\subsubsection{Retrieving A Query or Retrieving an Answer?}\label{retrieval-methods}

We believe the answer-based approach is more effective due to two reasons. (1) The query-based retrieval requires a query-to-evidence representation mapping. The answers are usually more semantically similar to the evidence, which avoids the mapping process. (2) A query is often very brief, while an answer containing the legal provision and rationale is more informative. In this sub-section, we analyze the retrieval component, including the apple-to-apple comparisons between the query-based and answer-based performance and the influence of answer quality for the answer-based retrieval.

We ordered the top-similar law clauses in each retrieval and evaluated the recall in top-$k$, indicating whether the ground-truth law clause appears in the top-$k$ retrieved law clauses. As shown in Figure~\ref{fig: retrieval}, the top-$1$ retrieved law clause based on the answer competes with the top-$5$ law clauses based on the query, and the answer-based retrieval beats the query-based retrieval with a large margin for all $k$. This confirms our first reason that the draft answer contains much more information than the query for retrieval, indicating that LLMs can be intrinsic retrievers.

We further compare the query-based and answer-based retrieval on a public \textbf{Similar Case Retrieval} task. This task aims to select similar legal judgments based on the query from the candidates with a query the case brief given. As shown in Table~\ref{table: similar-case-retrieval}, we compare two setups: 1) using the original query to retrieve, 2) using the legal 7B LLM to complete a whole legal judgment document given the brief query, and then retrieving. We follow the original task repository for the other settings\footnote{\url{https://github.com/myx666/LeCaRD}}. The results show that on each metric, the answer-based retrieval works better, indicating that using the generated answer provides a more robust retrieval.
\begin{table}[t!]
    \centering
    \resizebox{1.0\linewidth}{!}{
    \begin{tabular}{l|c|c|c}
    \toprule
        Setup&Precision@$5$& Precision@$10$&MAP\\
        \hline
        Query-based &42.1& 42.0& 47.8\\
        \hline
        Answer-based &\textbf{45.2} (+3.1)& \textbf{42.1} (+0.1)& \textbf{49.5} (+1.7)\\
        \bottomrule
    \end{tabular}
    }

\caption{\textbf{Results of two retrieval setups on the \textbf{Similar Case Retrieval} dataset}. } 
\label{table: similar-case-retrieval}
\end{table}

\begin{figure*}[t!]
    \centering
    \scalebox{0.82}{
    \includegraphics[width=\linewidth]{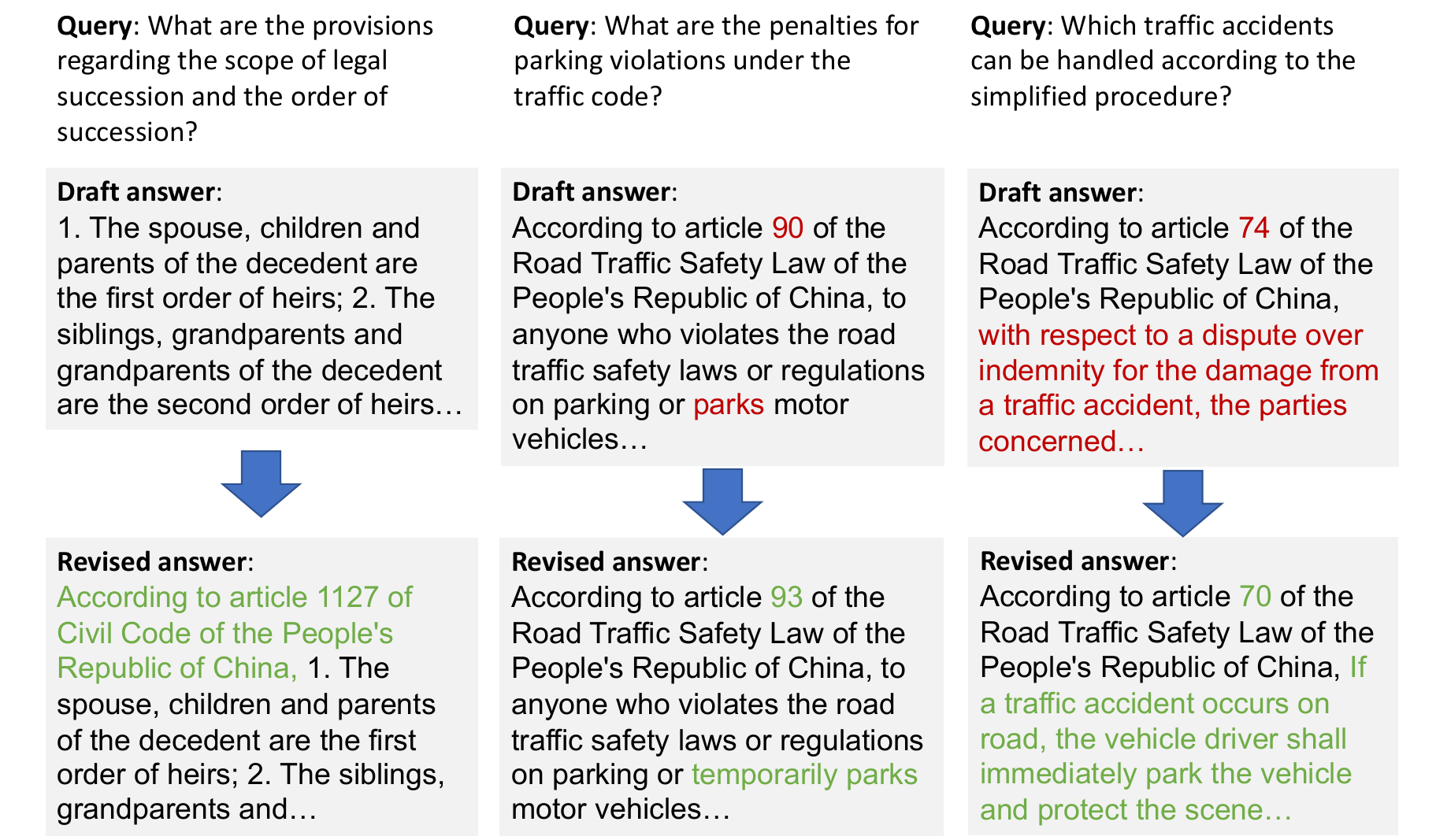}}
    \caption{\textbf{Case studies on the GPT-4 revision}. The examples are translated from Chinese to English for the demonstration purpose.}
    \label{fig: revision}
\end{figure*}

\subsubsection{Does the Quality of Answer Matter for Answer-Based Retrieval?}\label{similar-case-retrieval}
It's an intuitive thought that the quality of answers will significantly impact the outcome of answer-based retrieval. Therefore, we compare the retrieval using the answers of GPT-4 and the 7B legal LLM. We change the contents in retrieval for our proposed adapt-retrieve-revise method. As shown in Figure~\ref{fig: retrieval-results}, by comparing query-based and GPT-4's answer-based retrievals, we find that the answer-based retrieval fails on three datasets (LCR, CP, LegalQA), indicating that the lack of domain knowledge in the GPT-4 responses leads to a more noisy retrieval, which even hurts the performance of the draft answer (LCR, CP, LegalQA). Meanwhile, after domain adapting, our 7B legal LLM provides robust answers in retrieval and leads to the best performances, indicating that the learned Chinese legal domain knowledge improves our answer-based retrieval.

\subsection{Case Analysis of the Improvements after the GPT-4 Revision}
We conclude the improvements by GPT-4 in three aspects as shown in Figure~\ref{fig: revision}:
\begin{itemize}
  \item \textbf{Adding law clauses for reference}: Sometimes, the 7B legal LLM only provides a fluent response without following the input instructions to provide the key information of the referred law name and clause index. In this case, the faithfulness of the answer remains unchecked for the users. However, after the revision, each answer is equipped with the referred law clause, which makes it easier to check the accuracy of the responses.
  \item \textbf{Revising hallucinations in the evidence}: even the domain-adapted LLM can provide evidence from its learned legal knowledge; the hallucination remains to some degree, such as the wrong clause index, even the law name and rationale are roughly correct. Since the rationale content is accurate, the answer-based retrieval will search for the correct evidence, and the revision by GPT-4 will solve the hallucination to produce a more robust response.
  \item \textbf{Choosing correct evidence}: 
  In a more significant scenario, even though the 7B legal model's answers might contain partial hallucinatory content, the retrieval component can still possibly identify correct evidence through the partially correct descriptions in the rationale generation. During the revision stage, GPT-4 could assess the correct evidence, leading to the generation of correct answers.
\end{itemize}

\begin{figure}[t!]
    \centering
    \scalebox{1.0}{
    \includegraphics[width=\linewidth]{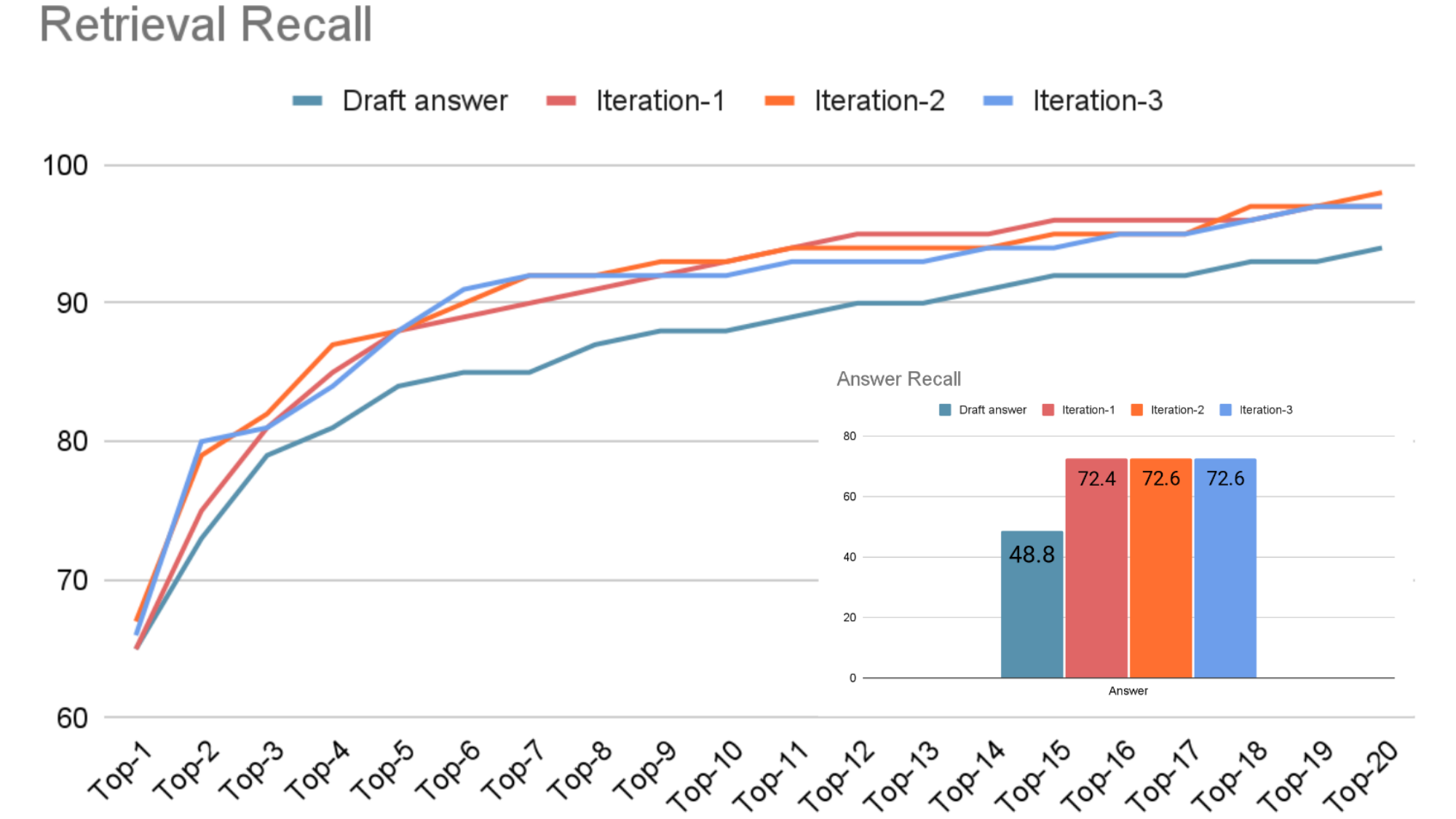}}
    \caption{\textbf{Comparison of iterations on LegalQA dataset.}}
    \label{fig: iteration}
\end{figure}

\subsection{Does the iteration make the generation better?}
Since our method provides more accurate responses than the original response from the domain LLM, one question is whether this procedure can be iterated to improve the responses. We can use the revised response to retrieve related evidence and further improve the response. To verify this probability, we iteratively test on the LegalQA dataset. As the result is shown in Figure~\ref{fig: iteration}, during the iteration, the retrieval recall does not show consistent improvements compared with the first revision, leading to the performance nearly unchanged.

\section{Related Work}
\subsection{Chinese Legal Domain Tasks}
The rapid advancements in LLMs have significantly impacted various domains, including the legal industry. This gives rise to the occurrence of legal datasets, such as the Challenge of AI in Law (CAILf)\footnote{ \url{http://cail.cipsc.org.cn/index.html}}, LeCaRD~\citep{DBLP:conf/sigir/MaSW000M21}, JEC-QA~\citep{DBLP:conf/aaai/ZhongXTZ0S20} and EQUALS~\citep{DBLP:conf/icail/ChenYZZSLS23}. These datasets cover document classification, summarization, question answering, information extraction, similar document retrieval, and other popular NLP tasks in the Chinese legal domain. To the best of our knowledge, this paper is the first work to exam the zero-shot performances on these legal datasets.
\subsection{Chinese Legal LLMs}
As for Chinese legal LLMs, recent work utilizes a paradigm of continual learning in the legal domain, and a substantial number of instruction fine-tuning datasets were constructed to augment the proficiency in rendering legal advice. Particularly, the series of LaWGPT~\citep{LawGPT} has been developed by leveraging foundational models such as Chinese-LLaMA-7B~\citep{cui2023efficient}, ChatGLM~\citep{du-etal-2022-glm}, and Chinese-alpaca-plus-7B~\citep{cui2023efficient}.
Lawyer LLaMa~\citep{huang2023lawyer} base on the more advanced Chinese-LLaMa-13B~\citep{cui2023efficient},  On the other hand, LexiLaw~\citep{LexiLaw}, built on the foundation of ChatGLM-6B~\citep{du-etal-2022-glm}, underwent training through the application of three distinct methods, namely LoRA~\citep{DBLP:conf/iclr/HuSWALWWC22}, P-tuning~\citep{DBLP:journals/corr/abs-2110-07602}, and fine-tuning. 
Furthermore, Chatlaw~\citep{cui2023chatlaw} received training based on both Ziya-LLaMA-13B-v1~\citep{Fengshenbang-lm} and Anima-33B~\citep{Anima}. DISC-LawLLM~\cite{yue2023disclawllm} adopted legal syllogism prompting strategies to construct supervised fine-tuning datasets and fine-tune LLMs with legal reasoning capability. 
A primary reason hindering us from utilizing such existing models is that they have often been trained on those publicly legal tasks already. Therefore the zero-shot capabilities can not be truly reflected. We thus continue training the general Baichuan 7B on legal data by ourselves.

\subsection{Retrieval-augmented Inference}
In scenarios where language models are confronted with tasks necessitating an infusion of external knowledge, a retriever mechanism can be used to provide evidence. The Retrieval-Augmented Generation (RAG)~\citep{DBLP:conf/nips/LewisPPPKGKLYR020} system incorporates a BERT-based~\citep{DBLP:conf/naacl/DevlinCLT19} Document Retrieval Process (DRP) and utilizes BART~\citep{DBLP:conf/acl/LewisLGGMLSZ20} for answer generation. Analogously, the EMDR2~\citep{DBLP:conf/iclr/0002IWXJ000023} employs the expectation-maximization algorithm to account for multiple retrieved documents.  The Atlas~\citep{izacard2022atlas} builds upon the EMDR2 framework, and by synergistically training the retriever and reader components, it demonstrates few-shot learning capabilities commensurate with the 540B PalM~\citep{chowdhery2022palm}. RETRO~\citep{pmlr-v162-borgeaud22a} benefits from retrieval mechanisms on expansive corpora during its pre-training phase and exhibits performance in close alignment with those of GPT-3~\citep{DBLP:conf/nips/BrownMRSKDNSSAA20}.
\section{Conclusions}
\label{sec:con}
In this paper, we reformulate the zero-shot domain content generation of large language models as an adapt-retrieve-revise procedure. This approach combines the merits of efficient continual training of a smaller 7B LLM for domain adaptation, robustly retrieving the supporting evidence from an external knowledge base, and effectively leveraging the evidence-assessing and revision capabilities of GPT-4. Our proposal substantially enhances the zero-shot performance on the Chinese legal tasks. 

\section{Limitations}
While this paper manages to validate the effectiveness of the proposal in the Chinese legal domain, the adapt-retrieve-revise method itself is a flexible framework, which is expected to be adapted to a wide range of domains. We leave it as future work. Due to the substantial costs of the GPT-4 API, we could only sample a subset of test data during the evaluation. Resolving the trade-off between the growing experimental costs and the validity of evaluation remains a challenge for the GPT-4 research in the future.

\bibliography{custom}
\clearpage
\appendix

\section{Appendix}
\label{sec:appendix}
\subsection{Legal Instruction Tuning}\label{legal sft}
We build our legal instruction dataset by human experts. Due to privacy concerns, we are not allowed to disclose the annotated instruction data. However, we will release the instruction annotation guideline along with our 7B legal LLM. We show a template with a toy example below.
\begin{itemize}
  \item \textbf{Due to the Article $x$ in the law $y$}: [the corresponding content in the law]
  \item \textbf{Considering the fact that} [the fact]
  \item \textbf{The judgment is} [the conclusion]
\end{itemize}
A toy example could be: 
\begin{itemize}
    \item \textbf{Due to the article 91 of the Road Traffic Safety Law of the People's Republic of China: }[Whoever drives a motor vehicle after drinking alcohol shall be imposed upon the penalty of temporary seizure of his motor vehicle driving license for not less than 1 month but not more than 3 months, and be imposed upon a fine of not less than 200 Yuan but not more than 500 Yuan as well; whoever drives a motor vehicle when he is drunk shall be restricted by the traffic administrative department of the public security organ until he becomes sober, be detained for not more than 15 days, be imposed upon the penalty of temporary seizure of his motor vehicle driving license for not less than 3 months but not more than 6 months, and be imposed upon a fine of not less than 500 Yuan but not more than 2000 Yuan as well. Whoever drives a commercial operating motor vehicle after drinking alcohol shall be imposed upon the penalty of temporary seizure of his motor vehicle driving license for 3 months, and be imposed upon a fine of 500 Yuan as well; whoever drives a commercial operating motor vehicle when he is drunk shall be restricted by the traffic administrative department of the public security organ until he becomes sober, be detained for not more than 15 days, be imposed upon the penalty of temporary seizure of his motor vehicle driving license for 6 months, be imposed upon a fine of 2000 Yuan as well. Where anyone is penalized for twice or more within one year due to his drunken driving as prescribed in the preceding two paragraphs, his motor vehicle driving license shall be canceled, and he shall not drive a commercial operating motor vehicle within 5 years.]
\item \textbf{Considering the fact that} [the man was riding a motorbike when drunk.] 
\item \textbf{The judgment is} [to be restricted by the traffic administrative department of the public security organ until he becomes sober, be detained for not more than 15 days, be imposed upon the penalty of temporary seizure of his motor vehicle driving license for not less than 3 months but not more than 6 months, and be imposed upon a fine of not less than 500 Yuan but not more than 2000 Yuan as well.]
\end{itemize}

\subsection{Retrieval Modules}
\label{retrieval comparison}
we leveraged multilingual E5-large, which is the SOTA family of text embeddings, which has been reported to outperform BM25, Contriever~\citep{izacard2022unsupervised}, and GPT embeddings~\citep{gptembeddings}. Since the improvements of our method are consistent and substantial (+33.3\% vs vanilla, +15.4\%/23.9\% vs retrieval baselines), we believe these gaps have shown sufficient robustness in our proposal.

However, we agree that including more established retrieval modules can enhance the robustness of our findings. Therefore, we added extra experiments and compared them with the current SOTA Chinese retrieval module CoROM following~\citep{qiu-etal-2022-dureader} on the LegalQA dataset, the same setting as in Section~\ref{retrieval modules}. Table~\ref{corom} shows the comparison results. We find that: (1) multilingual e5-large has a competitive performance with CoROM on query-based retrieval and vastly outperforms CoROM on answer-based retrieval; (2) for both modules, the answer-based retrieval primarily improves the retrieval quality than the query-based setting.
\begin{table}[ht!]
    \centering
    \resizebox{0.9\linewidth}{!}{
    \begin{tabular}{l|c|c|c|c}
    \toprule
        Retriever&Retrieval& Top-1&Top-5 & Top-10\\
        \hline
        Multilingual E5-large  &Query-based	& 45.8& 61.3 & 70.5\\
        \hline
        Multilingual E5-large  &Answer-based	& 65.3& 84.5 & 88.5\\
        \hline
        CoROM  &Query-based	& 47.5& 60.8 & 71.5\\
        \hline
        CoROM  &Answer-based	& 58.8& 72.5 & 80.5\\
        \bottomrule
    \end{tabular}
    }

\caption{\textbf{Results of different retrieval seteps}. } 
\label{corom}
\end{table}

\end{document}